\documentclass{article}
\usepackage{spconf,amsmath,epsfig}
\usepackage{spconf,amsmath,graphicx}
\usepackage[utf8]{inputenc}
\usepackage{graphicx}
\usepackage{verbatim}
\usepackage{array}
\usepackage{multirow}
\usepackage{pifont}
\usepackage{amssymb}
\usepackage{fontenc}
\usepackage{xcolor}
\usepackage{colortbl}
\usepackage{textcomp}
\usepackage{url}
\usepackage[linesnumbered,lined,boxed,commentsnumbered,ruled,vlined]{algorithm2e}

\let\OLDthebibliography\thebibliography
\renewcommand\thebibliography[1]{
  \OLDthebibliography{#1}
  \setlength{\parskip}{0pt}
  \setlength{\itemsep}{0pt plus 0.3ex}
}

\begin{document}\sloppy

% Example definitions.
% --------------------
\def\x{{\mathbf x}}
\def\L{{\cal L}}

% Title.
% ------
\title{Tabular Structure Detection from Document Images for Resource Constrained Devices Using A Row Based Similarity Measure}
%
% Single address.
% ---------------
\name{Soumyadeep Dey$^{*1}$, Jayanta Mukhopadhyay$^2$, Shamik Sural$^2$}
%Address and e-mail should NOT be added in the submission paper. They should be present only in the camera ready paper. 
\address{$^1$ Microsoft India, $^2$ Indian Institute of Technology Kharagpur \\ $^1$desoumya@microsoft.com, $^2$\{jay,shamik\}@cse.iitkgp.ernet.in}

\maketitle

\begin{abstract}

Tabular structures are used to present crucial information in a structured and crisp manner. Detection of such regions is of great importance for proper understanding of a document. Tabular structures can be of various layouts and types. Therefore, detection of these regions is a hard problem. Most of the existing techniques detect tables from a document image by using prior knowledge of the structures of the tables. However, these methods are not applicable for generalized tabular structures. In this work, we propose a similarity measure to find similarities between pairs of rows in a tabular structure. This similarity measure is utilized to identify a tabular region. Since the tabular regions are detected exploiting the similarities among all rows, the method is inherently independent of layouts of the tabular regions present in the training data. Moreover, the proposed similarity measure can be used to identify tabular regions without using large sets of parameters associated with recent deep learning based methods. Thus, the proposed method can easily be used with resource constrained devices such as mobile devices without much of an overhead.
\footnote{$^*$corresponding author. This work was done while the corresponding author was affiliated to Indian Institute of Technology Kharagpur}
\end{abstract}
\section{Introduction}
\label{sec:intro}

\begin{figure}[ht]
   \centering
 \begin{tabular}{cc} 
  \fbox{\includegraphics[width=.2\textwidth]{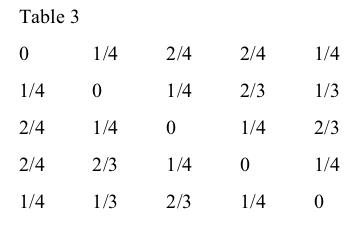}} &
  \fbox{\includegraphics[width=.2\textwidth]{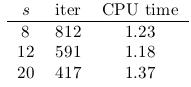}} \\  
 \end{tabular}
 \begin{tabular}{cc}
  \fbox{\includegraphics[width=.2\textwidth]{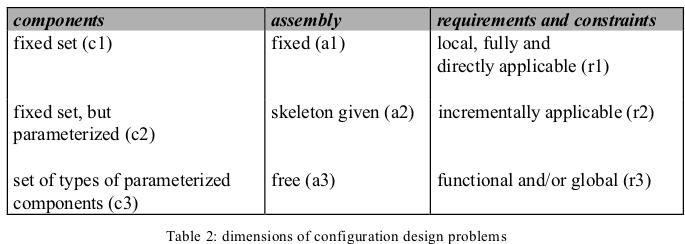}} & 
  \fbox{\includegraphics[width=.2\textwidth]{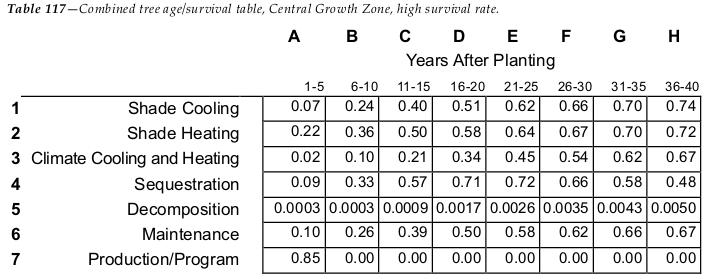}} \\
 \end{tabular}
 \caption{Sample Tabular Structures}
  \label{fig:sample_table_structure}
 \end{figure}

A large number of documents are produced everyday as scanned images. 
These documents disseminate knowledge and information in various forms, such as text, graphics, tables, etc. 
Tabular regions in a document are used to represent structural and functional information and play an effective role in fast understanding and summarizing the facts present in a document.
%Therefore, automatic detection of a tabular region in a document is an important task in document analysis. 
%Tables in a document may present in various form with variety of information within it. 
Structure of a tabular region varies from one document region to the other. 
Typical examples of various types of tabular structures present in different documents are shown in Fig.~\ref{fig:sample_table_structure}.
Since the tabular structures can be of diverse formats, automatic detection of these regions is quite challenging. 
Consequently, developing a generic tabular region detection algorithm handling the diverse structures is a nontrivial task.

Several work have been reported by researchers for detection and recognition of tables from document images. A brief survey of various table detection and recognition methods can be found in~\cite{Zanibbi2004}. 
These table detection algorithms are mostly constrained 
by the underlying assumption of certain formats or layouts of table structures~\cite{FangTable:2011}, \cite{Shafait_Table:2010}.  
One of the earliest systems for spotting and extracting table structures 
from document images is \emph{T-Recs}~\cite{Kieninger2001ApplyingTT}.
This system is limited to the extraction of words from a document, and its performance deteriorates in presence of multi-column documents. 
Mandal~\emph{et al.}~\cite{Mandal2006} propose a table detection algorithm
by analyzing gaps between the words in a text line. 
A table detection algorithm is proposed by Fang~\emph{et al.}~\cite{FangTable:2011} by analyzing both visual separator and table structure of a document.  
In recent years, researchers have tried to develop learning based table detection 
algorithms~\cite{AnhTable:2015}, \cite{Table_Rashid_ICDAR17}.  
A column and row line separator based table detection technique using SVM~\cite{libSVM} is proposed in~\cite{KasarTable:2013}. 
%In this method, vertical and horizontal lines are extracted, and are represented with $26$ dimensional feature vectors.
% In this method, a vertical and horizontal lines are extracted, 
% and each extracted line is represented with $26$ dimensional feature vector. 
%Later, SVM~\cite{libSVM} is used to detect whether the rule line is present in a table. 
%This method fails to detect a table in absence of rule lines. 
%Recently, deep-leaning based tabled detection methods have been reported in literature. 
Hao~\emph{et al.} ~\cite{Table_DAS16} propose a CNN~\cite{Goodfellow_DL_2016} based table detection technique. 
Another deep learning based table detection technique using Faster R-CNN~\cite{Ren:2015:FRCNN} is reported in~\cite{Table_Gilani_ICDAR17}. 
%An Euclidean distance based distance transform is applied on the input image. 
%This distance transformed image is then fed to Faster R-CNN~\cite{Ren:2015:FRCNN} to detect table regions.
Table detection can also be achieved through learning based 
generic document segmentation techniques like~\cite{Bouguelia_ICIP13}, \cite{wabdalmageed-08}, \cite{sdey_ijdar16}. 
The major limitation of learning based methods is that they are highly dependent on the training data.

\begin{figure*}[ht!]
 \centering
\fbox{\includegraphics[width=.85\textwidth]{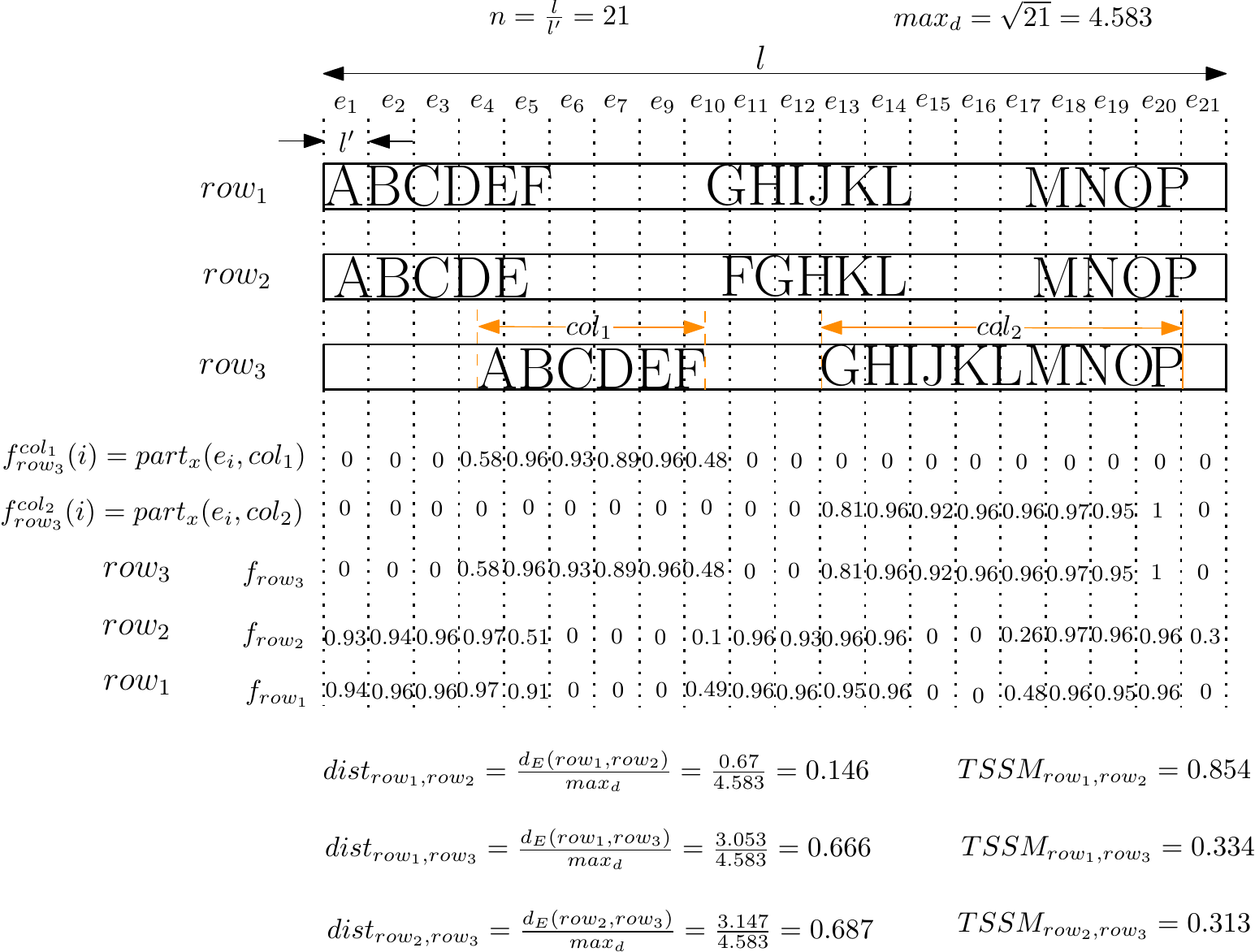}}
\caption{Sample Example showing Computation of $TSSM$}
 \label{fig:TSSM_example}
\end{figure*}

However, these table detection algorithms are mostly limited 
by the underlying assumption of certain formats or layouts of table structures~\cite{FangTable:2011}, \cite{Shafait_Table:2010}, \cite{Wang_icdar01}. 
Moreover, learning based algorithms~\cite{AnhTable:2015} \cite{Table_Rashid_ICDAR17} \cite{SilvaTable:2009} require a huge amount of training data for a satisfactory performance. 
In absence of such training data, the performance of learning based methods deteriorate. 
Further, even in presence of training data,a deep learning based methods learn and store millions of parameters to achieve a high performance. Therefore, attaining the desire degree of performance in a resource constraint device is quite challenging and difficult.

To the best of our knowledge, the existing techniques are limited to the detection of tables present in a document. They ignore other tabular regions like the legends present within a figure, etc. 
In view of limitations of the existing techniques discussed above, 
in this paper, a novel tabular structure detection technique is proposed. 
For detection of tabular structures a new similarity measure for finding 
similarities among a pair of rows within a tabular region is also introduced in this paper. 
This method is developed using the fact that tabular regions are used to represent structured and organized information, 
and rows within a tabular regions are used to represent those information. 
The proposed method can identify tabular regions of various layouts on resource constrained devices with same efficiency as on high end devices.

The technique for tabular region detection primarily consists of two steps. 
Initially from an input image, candidate tabular regions are detected. 
Then, tabular regions are detected based on the fact 
that, most of the rows in a tabular region are structurally similar and 
they are meant to represent similar information for various items in a structured way. 
Using this fact, the proposed similarity measure is used to detect structural similarities among the rows of a tabular region. 
A tabular region is detected based on the similarities among its rows.

 The rest of the paper is organized as follows. Section \ref{sec:SM} describes the proposed similarity measure. In Section \ref{sec:method}, the overall workflow of the tabular region detection algorithm is presented. Experimental results are presented and discussed in Section \ref{sec:Experiments} with Section \ref{sec:colclusion} concluding the paper.

\section{Tabular Structural Similarity Measure (TSSM)}
\label{sec:SM}
In this section, a new similarity measure is proposed which we name as Tabular Structural Similarity Measure (TSSM). 
This measure is used to detect structural similarities between two elements of a tabular region. 
These elements are the basic building blocks of a tabular region and can also be considered as rows of the region. In the rest of the paper, we shall use the terms elements and rows interchangeably. Here, it is assumed that these elements are of same \emph{length}, say $l$. 
For an element $e$ of a tabular region, $l$ represents the width of its rectangular cover 
and is computed using the following expression:\\
$len(box) = BR(box).x - TL(box).x$. Here, $BR(box).x$ and $TL(box).x$ respectively represent 
the $x$ coordinates of the bottom-right and top-left corners of a rectangular box.  {A tabular region element $e$ is divided into $n$ vertical blocks, say $e_{1}, e_{2}, \dots, e_{n}$, where $n = \frac{l}{l'}$, and $l'$ is the length of a primitive sub-element (for example average character width, considering characters to be the primitive sub-elements).  
An element $e$ is also divided into a set of \emph{horizontal regions} $H$.
These \emph{horizontal regions} are the regions which are separated from each other with a white space (gap) greater than a threshold $th_w$ in the horizontal direction.  
Each $col \in H$ of an element $e$ is represented using an $n$-dimensional feature $f^{col}_{e}$, 
where the $i^{th}$ feature ($f^{col}_{e}(i)$) of $f^{col}_{e}$ is the fraction of the $i^{th}$ part of the element 
that is horizontally covered by $col$. 
The horizontal part ($part_x(u,v)$) of an element $u$ covered by an element $v$ is computed according to Eq.~\ref{eq:hpart}.  
Here, operations $min(a,b)$ and $max(a,b)$, respectively, compute the minimum and the maximum of the two numbers $a$ and $b$.
\begin{small}
\begin{flalign}
\label{eq:hpart}
part_x(u,v)\ =\ \frac{M - N}{len(u)} \ \ \ \ \ \ \ \ \ \ \ \ \ \  \textrm{where,} \nonumber \\
M = min(BR(u).x,BR(v).x), \nonumber \\ 
N = max(TL(u).x,TL(v).x), and \nonumber\\
part_x(u,v) = \left\{\begin{array}{ll}
                           0 & \textrm{if} \ part_x(u,v) \leq 0 \\
                           part_x(u,v) & \textrm{otherwise} \\
                          \end{array}
                          \right.
\end{flalign}
\end{small}

$f^{col}_{e}(i)$ is computed using Eq.~\ref{eq:tab_ifeature}, where $e_i$ is the $i^{th}$ portion of the entire width of $e$. Here, $c$ is the sub-element of $col$ obtained by subdividing it with $th_w=0$.}
\begin{equation}
\label{eq:tab_ifeature}
f^{col}_{e}(i) = \sum_{\forall c \in col}{part_x(e_i,c)}
% f^{col}_{e}(i) = \frac{len(\textrm{part of $col$ in $e_i$})}{len(e_i)}
\end{equation}

%After finding $n$-dimensional feature for each column of a row, 
An element $e$ is represented with an $n$-dimensional feature $f_{e}$, where:
\begin{equation}
\label{eq:table_featurevec}
%f_{e} = \{f^{col}_{e}(1)f^{col}_{e}(2) \ldots f^{col}_{e}(n)\}
 f_{e} = \sum_{\forall col \in H}{f^{col}_{e}}
\end{equation}  
Based on Euclidean distance, the distance between two elements $u$ and $v$ ($d_E(u,v)$) is computed as 
% is used to find distance between any two rows of a given table region. 
%Euclidean distance between $v^{th}$ and $u^{th}$ row of a table is defined as:
\begin{equation}
\label{eq:tab_euclidean_dist}
d_E(u,v) = \sqrt[]{\sum_{i=1}^{n}{(f_{u}(i)-f_{v}(i))^{2}}}
\end{equation}
In Eq.~\ref{eq:tab_euclidean_dist}, $f_{u}(i)$ and $f_{v}(i)$ represent the $i^{th}$ feature of $f_u$ and $f_v$ respectively. 
As per the definition of $part_x(u,v)$ in Eq.~\ref{eq:hpart}, 
the value of $part_x(u,v)$ lies in the interval [$0, 1$]. 
According to Eqs.~\ref{eq:hpart} and~\ref{eq:tab_ifeature}, 
the value of the $i^{th}$ feature of the vector $f^{col}_{e}$ also belongs to the interval [$0, 1$]. 
$f^{col}_{e}(i)$ is $1$ if the $i^{th}$ part of $e$ is fully horizontally covered by its subregion $col$, 
and $f^{col}_{e}(i)$ is $0$ if the $i^{th}$ part of $e$ is not covered by $col$. 
Eqs.~\ref{eq:tab_ifeature} and \ref{eq:table_featurevec} may form two extreme vectors: 
a vector with $n$ zeros ($\{00\ldots 0\}$) and a vector with $n$ ones ($\{11\ldots 1\}$). 
The vector with $n$ ones corresponds to an element $e$ whose all parts are horizontally covered by its subregions. 
An element $e$ is represented by the vector with $n$ zeros if $e$ has no subregions. 
Therefore, according to Eq.~\ref{eq:tab_euclidean_dist}, 
the maximum distance between any two elements is computed as $max_d=\sqrt{n}$. 
A normalized distance ($dist_{u,v}$) is used to represent the distance between the elements $u$ and $v$, 
and it is computed using Eq.~\ref{eq:table_dist}.
%where $dist_{u,v} = \frac{d_E(u,v)}{max_d}$. 
 \begin{equation}
 \label{eq:table_dist}
  dist_{u,v} = \frac{d_E(u,v)}{max_d}
 \end{equation}
A pair of elements is considered to be similar, if the similarity between them is above a pre-defined threshold $th_{sim}$. 
Here, the similarity between the elements $u$ and $v$ is referred to as $TSSM_{u,v}$ and is computed as in Eq.~\ref{eq:table_row_sim}.
%where $sim_{uv} = 1-dist_{uv}$. 
%Here, $th_{sim}$ is set as $0.85$. 
% is computed as in Eq.~\ref{eq:table_row_sim}. 
 \begin{equation}
 \label{eq:table_row_sim}
 TSSM_{u,v} = 1-dist_{u,v}
 \end{equation}
{ 
A typical example illustrating the computation of $TSSM$ is shown in Fig.~\ref{fig:TSSM_example}. 
In this example, characters are considered as primitive subelements. 
Length $l'$ is the average width of all the characters. 
The example shows three rows (elements) namely $row_1$, $row_2$, and $row_3$. 
Length of all the rows is the same and it is represented as $l$. 
Each row is sub-divided into $n$ parts, where $n=\frac{l}{l'}$. 
In this example $n=21$. 
Therefore, each row in this example is represented with a $21$-dimensional feature and $max_d=\sqrt{21}$. 
Considering $th_w$ = $10$, 
$row_3$ consists of two horizontal regions namely, $col_1$ and $col_2$. 
Two separate feature vectors are computed for $col_1$ and $col_2$. 
While computing $f_{row_k}^{col_j}(i)$, $th_w$ is selected as $0$. 
For $row_3$, two separate feature vectors corresponding to $col_1$ and $col_2$ are computed. 
The feature $f_{row_3}$ is computed as $f_{row_3} = f_{row_3}^{col_1} + f_{row_3}^{col_2}$. 
All these computations are shown in Fig.~\ref{fig:TSSM_example}. 
In this figure, black dotted lines represent the partition positions of the rows $row_1,row_2,row_3$. 
Any value of the feature $f_{row_3}^{col_1}$ at $i^{th}$ partition is represented with $f_{row_3}^{col_1}(i)$. 
For example, $f_{row_3}^{col_1}(10) = 0.48$. 
This value represents that $col_1$ of $row_3$ covers $48\%$ of the partition $e_{10}$. 
Using this procedure, $row_1$, $row_2$, and $row_3$ are represented with $f_{row_1}$, $f_{row_2}$, and $f_{row_3}$ features, respectively. 
After computing the features, $TSSM$ measures among the rows are computed. 
Computed $TSSM$ measures among the rows are also shown in Fig.~\ref{fig:TSSM_example}. 
From the pictorial representations of $row_1$, $row_2$, and $row_3$ it is observed that, 
$row_1$ and $row_2$ are similar, $row_1$ and $row_3$ are dissimilar, and $row_2$ and $row_3$ are dissimilar. 
These visual similarities and dissimilarities are also reflected through 
the respective $TSSM$ measures $TSSM_{row_1,row_2}$, $TSSM_{row_1,row_3}$, and $TSSM_{row_2,row_3}$. }

\section{Tabular region detection from document images}
\label{sec:method}

\begin{figure*}[ht!]
 \centering
\includegraphics[width=.9\textwidth]{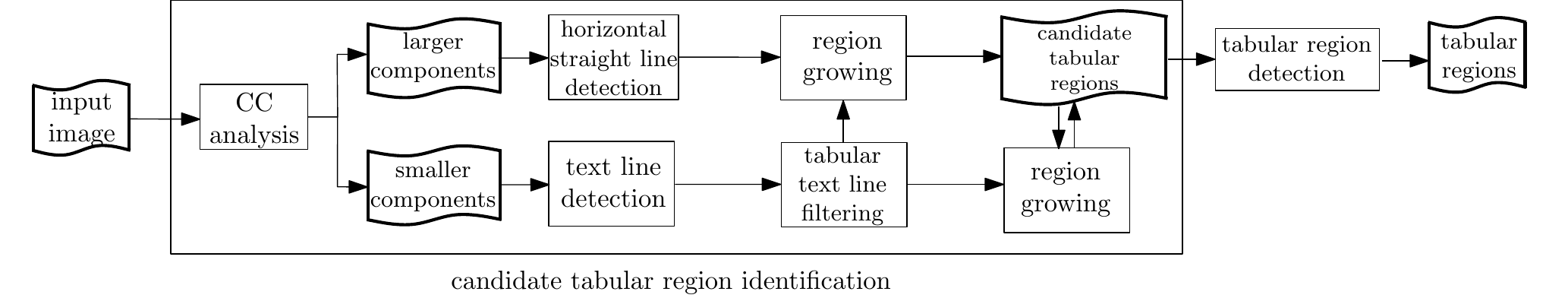}
% % TableFlow.pdf: 595x842 pixel, 72dpi, 20.99x29.70 cm, bb=0 0 595 842
\caption{Workflow of Proposed Tabular Region Detection Technique}
 \label{fig:table_workflow}
\end{figure*}

Overall workflow of the proposed tabular region detection method is shown in Fig.~\ref{fig:table_workflow}. 
For an input scanned document image, initially, a set of candidate tabular regions is identified.  
Then TSSM is used to determine the candidature of a candidate tabular region. 
Later, detected tabular regions are fine tuned to identify tables from a scanned document image using a few heuristic rules.

\section{Experimental results and discussion}
\label{sec:Experiments}
In this section, we describe the dataset and evaluation metrics used for experimentation and also present the experimental results.

\subsection{Dataset}
\label{subsec:dataset}

Experiments were carried out on the \emph{POD} dataset~\cite{POD17}. 
%The dataset consists of $2000$ images. 
This dataset was used for page object detection (\emph{POD}) competition of ICDAR 2017. 
The dataset consists of $2416$ images, out of which $1600$ images are meant for training and $816$ images are meant for testing. 
The dataset is prepared from $1500$ scientific papers from \emph{CiteSeer} with various layouts and regions like formulas, figures, and tables. 

For evaluation, \emph{ICDAR2009} segmentation dataset~\cite{ICDARSeg2009} is also used. 
This dataset consists of documents with complex layout with different varieties of regions. 
In the dataset, there are regions like, graphics, image, text, chart, mathematical equations, separators, table, etc.

\subsection{Evaluation metrics}
\label{subsec:evalmetrics}

The proposed method is evaluated in two stages. 
The method is evaluated as a table detection algorithm. 
For evaluation of the proposed table detection algorithm, $precision$, $recall$, and $F_1score$ are computed,  
where, $precision=\frac{tp}{tp+fp}$, $recall=\frac{tp}{tp+fn}$, 
and $F_1score=2 \times \frac{precision*recall}{precision+recall}$. 
Here, $tp$, $fp$, and $fn$ are computed using intersection over union (\emph{IOU}). 
$IOU$ helps to identify whether the table area detected by the algorithm matches with the groundtruth data. 
Let $A_d$ denote the area of the detected table, and $A_g$ be the area of the groundtruth table, 
then $IOU$ is defined as in~\cite{POD17}.
\begin{equation}
\label{eq:IOU}
IOU = \frac{A_d\cap A_g}{A_d+A_g-(A_d\cap A_g)}
\end{equation}
In Eq.~\ref{eq:IOU}, $A_d\cap A_g$ denotes the area of intersection of the detected table with respect to its groundtruth area. 
A table is said to be correctly detected if the $IOU$ score of a detected table is above $0.8$.
%the predefined $IOU$ threshold, 
%the table is said to be correctly detected, and results in an increment in the $tp$ count. $tp+fp$ is the total number of tables detected by the algorithm from the test data, and
%$tp+fn$ is the total number of tables actually present in the test data. 
%In this paper, the proposed technique is evaluated using $IOU$ threshold value as $0.8$. 

\subsection{Experimental Results}
\label{subsec:results}
\begin{figure}[h!]
  \centering
\begin{tabular}{cc}
 \fbox{\includegraphics[width=.2\textwidth]{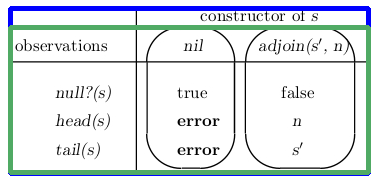}} &  
 \fbox{\includegraphics[width=.2\textwidth]{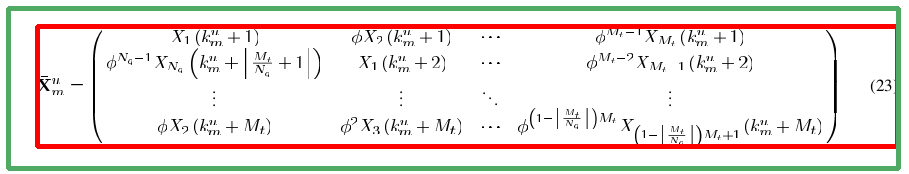}} \\  
 (a) & (b) \\
\end{tabular}
\begin{tabular}{cc}
 \fbox{\includegraphics[width=.2\textwidth]{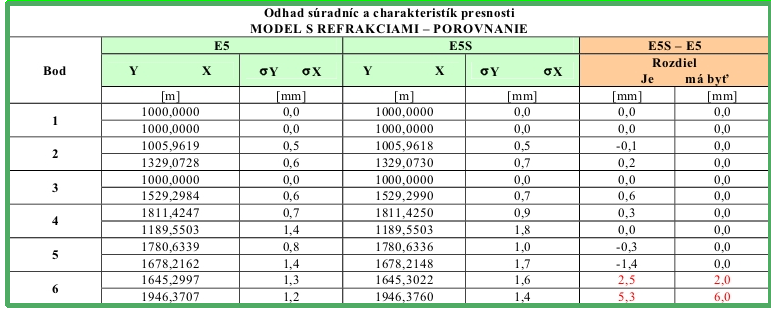}} &   
 \fbox{\includegraphics[width=.2\textwidth]{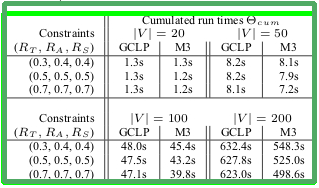}} \\
 (c) & (d) \\
\end{tabular}
\begin{tabular}{cc} 
 \fbox{\includegraphics[width=.2\textwidth]{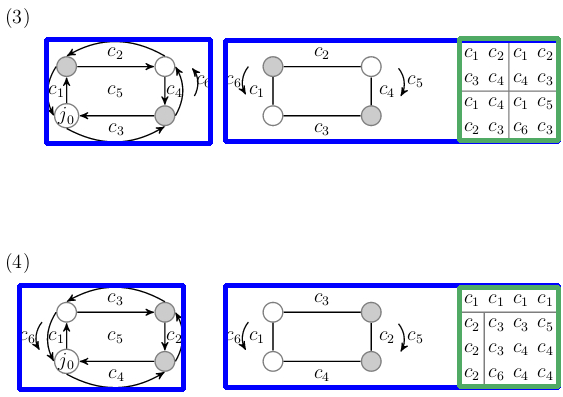}} & 
 \fbox{\includegraphics[width=.2\textwidth]{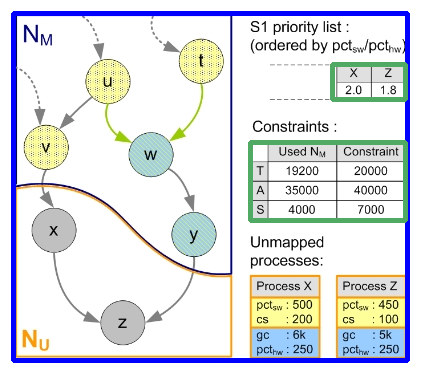}} \\
 (e) & (f) \\
\end{tabular}
\caption{Examples of Detected Tabular Regions from the \emph{POD} Dataset}
 \label{fig:wrong_result}
\end{figure}

The proposed method is evaluated on the datasets discussed in Sub-section~\ref{subsec:dataset}.
Tables~\ref{tab:tabularresults} and~\ref{tab:results} present the values of the 3 metrics discussed in Sub-section~\ref{subsec:evalmetrics} for tabular region detection. 
In these tables, we refer to our proposed method as TSSM table detector. 
A few examples of detected tabular structures by the TSSM method are shown in Fig.~\ref{fig:wrong_result}. 
%In Figs.~\ref{fig:wrong_result} and~\ref{fig:detected_tables}, 
In Fig.~\ref{fig:wrong_result}, 
groundtruth of the formula, table, and figure regions are represented with red, light green, and blue colored rectangles, respectively.
The tabular regions detected by the proposed algorithm are shown with deep green colored rectangles.

\begin{table}[h!]
\centering
\caption{Comparative results for table detection on POD dataset with statistical/ML based methods}
\label{tab:tabularresults}
\begin{tabular}{|c|c|c|c|}
\hline
method & $\textrm{precision}$ & $\textrm{recall}$ & $F_1score$ \\ \hline
\cellcolor[gray]{0.8}TSSM table detector & \cellcolor[gray]{0.8}$0.84$ & \cellcolor[gray]{0.8}$0.804$ & \cellcolor[gray]{0.8}$0.822$ \\ \hline
Mandal~\emph{et al.}~\cite{Mandal2006} & $<0.1$ & $<0.1$ & $<0.1$ \\ \hline
Almageed~\emph{et al.}~\cite{wabdalmageed-08} & $<0.2$ & $<0.2$ & $<0.2$ \\ \hline
Bouguelia~\emph{et al.}~\cite{Bouguelia_ICIP13} & $<0.1$ & $<0.1$ & $<0.1$ \\ \hline
Dey~\emph{et al.}~\cite{sdey_ijdar16} & $<0.3$ & $<0.3$ & $<0.3$ \\ \hline
\end{tabular}
\end{table}

The proposed tabular region detection technique is evaluated using the metrics defined in sub-section~\ref{subsec:evalmetrics}. 
The proposed method is compared with two different types of methods. 
In Table~\ref{tab:tabularresults}, the proposed method is compared with the methods which can be easily integrated in  resource constrained devices. 
In Table~\ref{tab:results}, the proposed method is compared with the top performing state-of-the-art deep learning based methods.

In Table~\ref{tab:tabularresults}, a table detection technique~\cite{Mandal2006} 
and three generic segmentation techniques~\cite{Bouguelia_ICIP13}, \cite{wabdalmageed-08} and \cite{sdey_ijdar16} are used for comparison. 
The comparative study of these methods is shown in the table. 
From the table, it is observed that the proposed method outperforms all the other methods in terms of tabular region detection. 
The method described in~\cite{Mandal2006} fails to detect a table from a document with heterogeneous arrangement of items, 
and consequently gives poor performance. 
A tabular region primarily consists of texts. 
In many cases, the learning based generic segmentation techniques~~\cite{Bouguelia_ICIP13}, \cite{wabdalmageed-08}, \cite{sdey_ijdar16} 
cannot distinguish between original text regions and table text regions.

The existing table detection methods only identify table regions from a document. 
On the other hand, the generic segmentation techniques cannot differentiate between texts in tabular regions and textual regions of a document. 
In contrast, our TSSM table detector tries to find structural similarities among the rows within a tabular structure. 
In the process, the method also detects tabular structures present within figures and formulas. 
Since this method detects tabular regions based on structural similarities among its rows, 
a region having structurally similar rows is detected as a tabular region.  

\begin{table}[h!]
\centering
\caption{Comparative results for table detection on POD dataset with deep learning based methods}
\label{tab:results}
\begin{tabular}{|c|c|c|c|}
\hline
methods & $precision$ & $recall$ & $F_1score$ \\ \hline
\cellcolor[gray]{0.8}TSSM table detector & \cellcolor[gray]{0.8}$0.84$ & \cellcolor[gray]{0.8}$0.804$ & \cellcolor[gray]{0.8}$0.822$ \\ \hline
NLPR-PAL & $0.958$ & $0.943$ & $0.951$ \\ \hline
FastDetectors & $0.879$ & $0.915$ & $0.896$ \\ \hline
Vislnt & $0.829$ & $0.823$ & $0.826$ \\ \hline
School of Software & $0.793$ & $0.798$ & $0.796$ \\ \hline
maitai-ee & $0.755$ & $0.798$ & $0.776$ \\ \hline
\end{tabular}
\end{table}

Comparative study of the proposed method with the top performing deep learning based methods on POD dataset is shown in Table~\ref{tab:results}. 
From this table, it is observed that, NLPR-PAL outperforms all other methods. 
However, TSSM table detector is ranked 3rd with respect to $precision$, 
and ranked as 4th with respect to $recall$.  
Among all the algorithms the $F_1score$ of the proposed method is the same as the Vislnt method and along with Vislnt ranked as 3rd.

The proposed tabular region detection algorithm is also evaluated on \emph{ICDAR2009} segmentation dataset~\cite{ICDARSeg2009}. On this dataset the proposed method achieves a $precision$ of $\mathbf{0.93}$, $recall$ of $\mathbf{1.0}$, and $F_1score$ of $\mathbf{0.96}$.

As already indicated, though the proposed method does not completely outperform the current popular deep-learning based methods, it achieves an overall good performance. Moreover, the deep-learning based methods that we have selected for comparison require a considerable amount of resources and cannot be easily executed in a resource-constrained environment. More specifically, these methods are not directly applicable for low-end devices where a huge amount of resource commitment is not possible. In contrast, our proposed method is quite appropriate for low-end devices and can be easily executed in such environment. This advantage coupled with the good overall performance makes TSSM table detector a better choice in comparison to similar algorithms applicable for low-end devices as can be observed from the Tables~\ref{tab:tabularresults}. 
The methods listed in Table~\ref{tab:tabularresults}, clearly suffer from the disadvantage of performance in terms of all the three metrics even though they are applicable for low-end devices. 
Moreover, the performance of our proposed method is not bound by any resource commitment and gives uniform results in any kind of computing environments.

It has been observed during the experiment that, a few false positive cases are identified from formula regions, where consecutive equations in new lines result in structurally similar rows.  
%and results in wrong detection of tables. 
One such falsely detected tabular region is shown in the Fig.~\ref{fig:wrong_result}~(b). 
Few other false positive cases are also shown in Fig~\ref{fig:wrong_result}~(a), (e), and~(f). 
Here, tabular regions are detected from a graphics regions. 
Though, in these cases, detected tabular regions are part of graphics, the proposed method is able to detect actual tables present within the graphics region. 
From the detected tabular regions shown in Fig.~\ref{fig:wrong_result}, 
it may be observed that our method is able to detect 
tabular structures from various regions of a document in addition to table regions.

\section{Conclusion}
\label{sec:colclusion}

A new similarity measure is proposed in this paper to detect tabular regions. 
This measure finds structural similarities among the elements or rows within a tabular region. 
Using this similarity measure, a region is detected as a tabular region. 
The proposed method is efficient and effective even for memory and computing power constrained devices. 
This method can effectively detect tabular structures from a document, 
but in some cases, other regions having consecutive structurally similar lines (rows) like formula regions, etc. 
are detected as tabular regions. 
The tabular region detection technique is further fine tuned for the detection of tables. 

% use section* for acknowledgement
% \section*{Acknowledgment}
% This work is funded by TCS research scholar program and Ministry of Communications \& Information Technology, Government of India; 
% Ref.: MCIT $11(19)/2010 - \textit{HCC}$ (TDIL) dt. $28-12-2010$. 

\bibliographystyle{IEEEbib}
%\bibliography{strings,refs}
\bibliography{references.bib}

\end{document}